# A Proposed Biomedical Data Policy Framework to Reduce Fragmentation, Improve Quality, and Incentivize Sharing in Indian Healthcare in the era of Artificial Intelligence and Digital Health

Nikhil Mehta[1], Sachin Gupta[2], Gouri RP Anand[3]

[1]Consultant Dermatology, Department of Telemedicine, Post Graduate Institute of Medical Sciences, Chandigarh, India

[2]Assistant Professor, Department of Dermatology, Amrita School of Medicine, Amrita Vishwa Vidyapeetham, Faridabad, Haryana, India

[3]Senior Resident, Department of Dermatology, All India Institute of Medical Sciences, Delhi, India

Corresponding Author: Nikhil Mehta, Consultant Dermatology, Department of Telemedicine, Post Graduate Institute of Medical Sciences, Chandigarh, India

nikhilmehtadermatology@gmail.com

## SUMMARY

India generates vast biomedical data through postgraduate research, government hospital services and audits, government schemes, private hospitals and their electronic medical record (EMR) systems, insurance programs and standalone clinics. Unfortunately, these resources remain fragmented across institutional silos and vendor-locked EMR systems. The fundamental bottleneck is not technological but economic and academic. There is a systemic misalignment of incentives that renders data sharing a high-risk, low-reward activity for individual researchers and institutions.

Until India's academic promotion criteria, institutional rankings, and funding mechanisms explicitly recognize and reward data curation as professional work, the nation's AI ambitions will remain constrained by fragmented, non-interoperable datasets. We propose a multi-layered incentive architecture integrating recognition of data papers in National Medical Commission (NMC) promotion criteria, incorporation of open data metrics into the National Institutional Ranking Framework (NIRF), adoption of Shapley Value-based revenue sharing in federated learning consortia, and establishment of institutional data stewardship as a mainstream professional role. Critical barriers to data sharing, including fear of data quality scrutiny, concerns about misinterpretation, and selective reporting bias, are addressed through mandatory data quality assessment, structured peer review, and academic credit for auditing roles. The proposed framework directly addresses regulatory constraints introduced by the Digital Personal Data Protection Act 2023 (DPDPA), while constructively engaging with the National Data

Sharing and Accessibility Policy (NDSAP), Biotech-PRIDE Guidelines, and the Anusandhan National Research Foundation (ANRF) guidelines.

**Manuscript**

**1. INTRODUCTION: THE PARADOX OF DATA ABUNDANCE**

India is a country with the highest population and expectedly, with the highest prevalence in absolute numbers for many diseases, and one of the highest number of doctors in absolute terms. The country generates enormous volumes of clinical data through postgraduate research and doctoral research, investigator-initiated research, government hospital services and audits, government schemes and funding, private hospitals and their electronic medical record (EMR) systems, insurance programs and standalone clinics. Despite this, the number and the size of registries, data sets, longitudinal cohorts, and biobanks from India pales in comparison with many smaller nations. The cost of this missed opportunity, in the age of data being the new gold, is massive. This problem needs to be tackled urgently, more so in the current era of digital health and artificial intelligence, when we have increasing ability to use massive quality data for meaningful outcomes.

It is not that the data does not exist. It is often fragmented, siloed, and of poor quality. Most institutional clinical data either remain on paper files, or locked within proprietary EMR systems with poor interoperability, expensive de-identification workflows, and no mechanisms for structured reuse. It is not unusual to see hundreds of papers and posters in various medical conferences, most of which are never published. Similarly, many postgraduate and doctoral theses lie waste in institutional libraries without ever being available for compilation with similar datasets. It is such a waste to have multiple small sample-sized studies from various centers reporting almost similar data on a disease without the numbers or degree of robustness required to be considered credible evidence by guideline groups, consensus societies, and Cochrane reviews. The potential for longitudinal, multicentric datasets of sufficient size and diversity to build disease registries, generate population-level epidemiological insights, or train robust artificial intelligence (AI) predictive models, remains unrealized. Consequently, we know significantly less about how the diseases affect the Indian population compared to the other populations.

What leads to this data fragmentation and poor quality? This paper discusses that India's data fragmentation is a consequence of misaligned incentives rather than technical incapacity, and proposes a policy framework to realign individual and institutional interests with data sharing and quality control for public health and research goals. While having extensive good quality data is beneficial for many use cases, it is imperative for artificial intelligence. A system that is able to support reliable AI development will also strengthen clinical research, disease registries, and health system learning more broadly.

## 2. REGULATORY AND POLICY LANDSCAPE AND COMPARISON WITH OTHER NATIONS

### 2.1 The good

Significant improvements have been made by Government of India recently to address the issue of data. The ICMR National Ethical Guidelines (2017, along with subsequent ICMR addenda and guidance documents) endorsed "broad consent" for biobanking and future research uses.[1–3] However, "broad consent" contradicts the DPDPA's stricter "specificity" requirement.[4] The National Data Sharing and Accessibility Policy (NDSAP 2012) mandated sharing of non-sensitive, public-funded research data but lacked enforcement mechanisms and allowable costs.[5,6] In 2014, DST and DBT released their Open Access Policy, which was still without an enforceable mandate. In 2022, when Science and Engineering Research Board (SERB) moved to a 100% digital portal, they added a mandatory field in the final progress report about a link or Deposit ID from a recognized repository. With the transition of SERB into the Anusandhan National Research Foundation (ANRF), the Deposit ID is now often required not just at the end of a project, but as part of the Data Management Plan (DMP) at the start.[7] The Biotech-PRIDE Guidelines (2023) established the Indian Biological Data Centre (IBDC) as a centralized repository, solving the infrastructure question but not the motivation problem.[8]

The ANRF Act 2023 mandates that the foundation maintains a "central repository for data", though incentive mechanisms to populate this repository remain underdeveloped.[7] Statutory authority is insufficient without accompanying recognition and funding mechanisms.

In India, data-sharing "punishment" is administrative rather than criminal: non-compliant researchers are blacklisted on internal "defaulter lists" by agencies like ICMR and DST. This triggers automatic rejection of future grants and institutional friction, as agencies may withhold department-wide overhead funds. Practically, this hits your budget and career by withholding the final 10% of grant funding and stalling promotion files until a valid Deposit ID and Completion Certificate are verified.

Instead of just "punishing" non-compliance, the government is now incentivizing sharing. Secure AI for Health Initiative (SAHI) framework, released by Government of India at the AI Summit 2026, mentions that researchers who share high-quality datasets with the AIKosh platform receive subsidized access to high-end GPUs at roughly ₹65 per hour.[9] Such compute is highly desirable for techniques requiring complex data processing, like single cell RNA sequencing. This is similar to the NIH's All of Us program incentivizes participation by offering researchers privileged access to cloud-based analytics tools, GPU clusters, and pre-installed algorithms.[10] India's National Supercomputing Mission operates powerful machines chronically underutilized in smaller research institutions, which can be leveraged for this.[11]

Various standards, frameworks, and terminologies exist to permit technical and semantic interoperability. Observational Medical Outcomes Partnership (OMOP) is a CDM which is a global standard developed by the Observational Health Data Sciences and Informatics (OHDSI) community.[12] Such CDMs allow mapping various data to standardized vocabularies, enabling large-scale analytics. Systematized Nomenclature of Medicine - Clinical Terms (SNOMED CT) is the most comprehensive, multilingual clinical healthcare terminology in the world.[13] While systems like ICD-10 are designed for high-level classification (like billing or mortality statistics), SNOMED CT is designed to capture the granular detail of a clinical encounter. Similarly, Logical Observation Identifiers Names and Codes (LOINC) is the standard terminology for health measurements, be it clinical observations/scores or laboratory parameters.[14] Fast Healthcare Interoperability Resources created by Health Level 7 International (HL7 FHIR) allow granular data sharing and exchange between different EHR software. [15]

HL7 FHIR and LOINC are adopted and mandated by ABDM, similar to Australian Health Data Evidence Network (AHDEN) and Singapore Health Data Hub (which used the HealthX Innovation Sandbox).[14,16–19] The National Health Authority (NHA) has released a specific FHIR Implementation Guide that defines exactly how patient profiles, diagnostic reports, and prescriptions must be structured. SAHI implementation requires OMOP usage and SNOMED CT/LOINC terminologies.

The UK Biobank model emphasizes prestige and return-of-results mandates, requiring researchers to return derived datasets within six months of publication.[10] Disease-specific national registries in India could adopt this model and gain NIRF prestige, though massive upfront government investment and strong institutional governance are required.

Canada emphasizes federated governance where federal, provincial, and private-sector organizations maintain data sovereignty while participating in coordinated commons.[20] Given India's federal structure, a federated model where states and central government maintain data sovereignty while participating in an ANRF-coordinated commons is politically and operationally viable. This is also supported by DPDPA 2023.[4]

**2.2 The bad**

The DPDPA 2023

While the DPDPA 2023 provides significant agency to patients on their own data, and rightly shifts from an intrinsic trust approach to proper written informed consent approach, its proper adoption in the Indian biomedical research ecosystem faces some friction and resistance. It is quite new, and some sections have no precedence for their interpretation.

Section 17(2)(b) carves out an exemption for processing personal data for "research, archiving or statistical purposes," provided data is not used to inform decisions about identifiable individuals and "processing is carried on in accordance with such standards as may be prescribed." The standards and prescribing authority have not been specified. If one wants to use legacy data for retrospective studies, using anonymized data from clinical records after prior ethical approval by a valid ethics committee should be sufficient standards to waive off individual consent requirement. However, the distinction between "pure" research (epidemiological analysis of anonymized case series) and "applied" AI development (training diagnostic algorithms for clinical deployment) is legally ambiguous.

In a realistic scenario, a researcher or an institute, acting as a data fiduciary, partners with a third-party contractual service provider for data processing, lacking the processing capacity itself. If a leak or utilization without consent happens at the processor's end, as per the interpretation of the law, the vicarious legal liability and the massive fines fall squarely on the data fiduciary that collected the data. The researcher or the institute may sue the third-party for violation of the contract (which can have an indemnity clause) or data processing agreement, but that entails a significant legal headache and time delay in Indian civil courts. The data fiduciary will have to undertake this exercise while simultaneously defending itself for a breach that was not its fault. This liability asymmetry and risk calculus strongly discourages participation. Compared to this, the Omnibus rule in the USA leads to a shared responsibility with a business associate agreement.[21] The third-party service providers or business associates are directly liable to the government for HIPAA (Health Insurance Portability and Accountability Act) breaches irrespective of what their contract with the covered entity (researcher/institute) states. Similarly, EU's GDPR (General Data Protection Regulation) too, processes have a direct statutory obligation to the European regulators.[22] So Indian researchers and institutions have a disproportionately greater liability, and less incentives for electronic data records and data sharing, compared to them. DPDPA could evolve to introduce carefully scoped obligations for large data holders to support public-interest research. Domain-specific data trusts could serve as intermediaries, negotiating Data Use Agreements and ensuring fair attribution.

**2.3 The ugly?**

While the policies and frameworks have excellent vision, but the implementation and execution need to be significantly stronger.

**3. STRUCTURAL BARRIERS TO DATA SHARING**

**3.1 Academic credit system**

Publication in indexed journals remains the primary currency of selection and academic career progression under National Medical Council (NMC) and University Grants Commission (UGC) rules and Academic Performance

Indicator (API) system.[23,24] Data curation and registry maintenance, the labor-intensive skilled work of collecting, coding, validating, and formatting data for secondary use, is invisible in these reward systems.

Faculty rationally prioritize manuscript writing over data curation, even if the manuscripts may not be novel. So due to misplaced priorities, the intellectual might of Indian academia prioritizes its time publishing non-novel case reports, or original articles involving poorly collected data in non-indexed predatory journals, without any relevance or practical utility. The editorship of a journal is a prestigious position, and legacy journals are often strongly gatekept among institutions or labs. So, even founders/editors of newer journals with good intentions, hoping to get indexed later, are sometimes forced to accept such manuscripts, as they might not get enough quality submissions. So, what we get are individual scattered publications of the same condition, often without the master sheet of the whole anonymized data, inaccessible for manual or AI compilation. Such data doesn't serve any other function except adding to the publication lists of the authors, and can have more authors than readers. Additionally, many academicians become part of 'citation circles', lending credibility to such studies. This leads to a data quality problem, expounded next.

## 3.2 Data quality

India has a significant data quality problem. The trust perception of studies coming from India is low. India frequently ranks in the top list of paper retractions. The matrices to judge performance are usually based on research quantities rather than the quality. A clinician could have documented cases hastily, with missing variables, inconsistent coding, or incomplete follow-up. The submitted excel sheets can be squeaky clean and result in similar results on independent statistical analysis, but the data entered itself can be fake or manipulated. This leads to problems in reproducibility of results in clinical settings and later studies. When comprehensive datasets are shared, these methodological gaps become visible to scrutinizers.

Even genuine researchers do not share data, fearing misinterpretation: that others analyzing their data using different statistical approaches or populations will draw conclusions contradicting the original authors' interpretations, damaging reputations and opening liability disputes.[25] These are rational fears. Without structured mechanisms for quality verification, auditing, and transparent attribution of findings to methodology, data sharing invites reputational risk.[26]

## 3.3 No mandated data sharing

Indian researchers treat data as a private asset rather than a public good. While research funded from agencies like ICMR and DST require data to be deposited as a part of the final report, this data is not checked manually. For those sponsored by other agencies/institutions or non-sponsored studies, there is no enforced mandate for public data sharing. The researchers argue that non-funded studies on their own initiative do not warrant public sharing. But the researchers of public institutions are able to collect such data or perform these self-initiated studies by the virtue of being employed by public funded institutions that provide them access to patients, samples, and ethics committees.

### 3.4 Authorship hierarchy and failure of due recognition

In large multicentric studies, peripheral hospitals are relegated to "group author" status with zero weight in NMC promotion criteria.[27] So, in a multi-centric study, a centre contributing lesser number of cases due to its peripheral geographical location may have its author listed only as a part of a consortia. It could have represented a diverse data subset which may have certain distinctions from the whole data set. But that's not a metric which is valued. The consequence is that smaller institutions opt out of collaborations, retaining data locally.

Ambiguity around data ownership also deters sharing. A researcher at a government institution might fear that sharing data with external consortia could lead to intellectual property being claimed by others or commercialized without acknowledgment or benefit-sharing. Even if she/he signs a memorandum of understanding with partner institutes, practically the enforcement of penalties for any breaches remains poor and drawn out, fraught with delaying tactics such as requesting repeated physical visits instead of online resolution, and communication with post rather than e-mail. There is a perception, not unfounded, that the bigger names in academia can get away with stealing data and credit because of their prestige, undue influence in academic circle, and proximity to power.

### 3.5 Lack of attribution and data provenance

India lacks a unified system for tracking micro-contributions to datasets. There is no national ledger recording "researcher contributed 50 dermoscopy images to Dataset B" with persistent identifiers. Without provenance infrastructure identifying who originated the data, downstream revenue-sharing, citation credit, or algorithmic valuation is technically impossible.

### 3.6 Manpower and infrastructure resource constraints

Data exists in heterogeneous forms: legacy paper records, disparate EMR systems, image files in varying compression standards, semi-structured notes. Converting this into a FAIR (Findable, Accessible, Interoperable, Reusable) dataset requires skilled labor. It involves de-identification, standardization to common data models (CDM), validation, metadata creation, and persistent identifier assignment.[28,29]

Conversion of retrospective clinic data to OMOP requires and manpower to extract, transform, and load, despite existence of open-source tools. These manpower and infrastructure requirements are typically not budgeted in research grants. Indian funding agencies have historically not included "data management and curation" as an allowable cost. In comparison, the US National Institutes of Health explicitly permits budgeting for data curators and repository fees.[30]

# 5. COMPREHENSIVE SOLUTIONS

## 5.1 Individual Academic Recognition: Data Papers and Revised Promotion Criteria

### 5.1.1 NMC Recognition of Data Papers

The NMC should amend its Teachers Eligibility Qualifications Regulations to recognize "Data Papers" as peer-reviewed articles describing dataset methods, metadata, validation protocols, and reuse applications.[31,32] These papers should be equivalent to original research articles for promotion purposes and published in venues like Scientific Data (Nature), Data in Brief (Elsevier), or domain journals with a persistent Digital Object Identifier (DOI) linking to a recognized repository (IBDC, ICMR portal, or international repositories like Zenodo). This immediately legitimizes data work as research output. It provides first-author credit upon data sharing, not years later when others publish analyses. Subsequent users must cite the Data Paper, increasing the original researcher's h-index.

Publication of data set of non-funded individual-led or postgraduate research in open domain in a time-bound manner should be a mandatory criterion at the ethical approval stage itself. What a tickle of the world of a new research project, every researcher should be made to sign a declaration that the data set of any prior research finished at least six months ago should have been published in approved databases in universal terminology. Double dipping should be permitted, for example, a researcher conducting a study on atopic dermatitis should be able to avail two authorship credits from the study: one immediately as a data paper, and later when the inference/results of the data is published. Perhaps a third time too if the data is used for AI-based analysis. This would encourage professors to put their post-graduate research data in the open domain, and to actively seek permission from patients, to allow contribution of such data for AI-based analysis.

### 5.1.2 NMC Clinical Registry Contributions

Indian academics aiming to start journals should chair and manage those specific domains in data registries instead. Registry management and coordination should attract as much prestige, or at least formal credit, as a journal editorship. The NMC should recognize verified contributions to national disease registries (minimum 50 fully annotated cases) as a valid research activity, equivalent to publishing case reports or observational studies. Academic systems are increasingly acknowledging that research data contributions warrant formal credit, similar to publications and citations.[33] Clinical registries are recognised as high-value research infrastructure that enable longitudinal and observational research across populations.[34] The registry must issue a certificate verifying data quality and adherence to standards. A National Dermatology Image and Registry Initiative could be established with predefined minimum datasets covering clinical diagnosis codes (ICD-11), lesion morphology, anatomical site mapping, dermoscopy or histopathology data, and treatment outcomes. Dermatologists contributing high-quality cases with standardized documentation would earn registry research credit.

## 5.2 Institutional Academic Recognition: UGC and NIRF

### 5.2.1 UGC Academic Performance Indicators

The UGC should introduce a new sub-category under "Research and Academic Contributions" specifically for "Digital Research Assets," with explicit scoring for datasets, software, and code. This aligns with the UGC's pivot towards outcome-based evaluation. In fields like Bioinformatics, Computer Science, and Clinical AI, the primary research output is often code, data, or models, not prose.[35]

### 5.2.2 Institutional Recognition: NIRF Integration

The Ministry of Education should introduce an "Open Science and Data Stewardship" parameter in the NIRF overall ranking with measurable weightage, including weighted publications (example 1.2x multiplier) for those with Data Availability Statements, data citation impact tracked via Scholix framework, and registration of datasets in IBDC/ICMR as "Registered Datasets.[27,36]

Institutions would be required to declare data assets and provide evidence of FAIR compliance, with third-party auditors verifying claims. This should lead to a "race to the top" for data where premier institutions establish Data Stewardship Offices.

The National Assessment and Accreditation Council (NAAC) and National Accreditation Board for Hospitals (NABH) should incorporate data sharing, registry participation, and AI governance as formal accreditation criteria. This should be incorporated in institutional Data Management Plans, participation in at least one data-sharing consortium, annual registry contribution audits, and Ethics Committee oversight of AI projects with explicit bias mitigation protocols.

## 5.3 Financial Mechanisms

### 5.3.1 Mandatory Data Management Budget Lines

All central government research grants (ICMR, ANRF, DBT, DST) above a certain amount, say 50 lakhs, must include dedicated budget lines for "Data Management and Sharing," including personnel (data curators, annotation specialists, example at 30-50 percent salary allocation), repository fees, infrastructure, and training. International funding agencies have explicitly recognised data stewardship as a legitimate and necessary research cost, permitting budgetary allocation for data curation staff, documentation, metadata creation, and repository fees.

Data Management Plans (DMPs) should become a mandatory component of grant proposals, reviewed and scored as part of scientific review. DMP completion should be a condition of grant continuation and final disbursement, consistent with established practices of major international funding programs.[10]

### 5.3.2 Shapley Value-Based Revenue Sharing

For large ANRF-funded multicentric AI projects, adopt Shapley Value algorithms to fairly distribute incentives based on each institution's marginal contribution to model performance.[37] Instead of assuming all data points are equally valuable, algorithms like SaFE (Shapley for Federated Learning using Ensembling) calculate how much each institution's data improves the final model's accuracy.[38]

A large tertiary centre contributing 100,000 dermatology images of common conditions and a smaller district hospital contributing 5,000 images but including 500 rare cases of epidermolysis bullosa would traditionally see allocation weighted 95 percent to volume. Shapley Value recognizes that the rare cases increase model utility by 25-40 percent, earning the district hospital a much larger share. This encourages smaller, specialized institutions to participate.

### 5.3.3 Blockchain-Based Consent and Royalty Systems

Implement blockchain-backed smart contracts on top of ABDM to automate transparent tracking of data use and royalty distribution. When a patient consents to data sharing for a specific AI project, this consent is recorded on a blockchain-backed registry with a unique transaction ID. If the resulting AI model is commercialized, smart contracts automatically trigger micro-payments. For example, a system can be Data Fiduciary (hospital) 60-70 percent, Data Principal (patient or advocacy group) 10-20 percent, and Data Trust (governance overhead) 10-20 percent. All parties can query the blockchain to see how their data is being used and what benefits have accrued. Blockchain and smart-contract architectures have been proposed and prototyped for medical data access control, consent auditing, and monetisation, demonstrating technical feasibility while also highlighting governance, privacy, and scalability considerations that must be addressed in any national implementation.[39–41]

## 5.4 Technical Infrastructure and Standards

### 5.4.1 Federated Learning Platform

Establish a national federated learning platform within ABDM or ICMR sandbox infrastructure, enabling multicentric AI development without centralizing raw data.[38] Each hospital trains a diagnostic AI model locally using only their data; only model weights (updated parameters) are sent to a central server, not raw images or patient data. Central servers aggregate updates using privacy-preserving techniques: averaging, differential privacy, homomorphic encryption. This addresses DPDPA compliance by never transferring raw patient data across institutional boundaries.

### 5.4.2 Domain-Specific Data Trusts

Establish specialty-specific data trusts under professional societies (for example, IADVL for dermatology). These trusts negotiate standardized Data Use Agreements, ensure fair attribution and co-authorship protocols, enforce obligations (return of AI tools to participating centres, local validation before deployment), and mediate public-private partnerships.

For example, a proposed IADVL Dermatology Data Trust would include: steering committee with representatives from teaching hospitals, private practices, patient advocacy groups, and ethicists; standardized, legally vetted DUA specifying academic-only use (unless explicit benefit-sharing); AI models returned to participating public hospitals at no cost within 12 months; contributors guaranteed co-authorship if ≥10 percent data contribution; and commercial licensing royalties distributed per Shapley Value.

### 5.5 Data Quality and Bias Mitigation

#### 5.5.1 Structured Peer Review and Data Auditing

A time-bound formal "Data Audit and Peer Review" process for datasets should be established before they are deposited in central repositories or used in AI projects. This process mirrors traditional manuscript peer review but focuses on data quality, completeness, and methodology rather than conclusions. Not just the results of the data but even the acquisition should be tested. For example, calling (with prior consent) a fixed proportion of patients (around 5-10%) in a study to verify the inclusion and correctness of epidemiological parameters by independent external (separate department or institution) reviewers, and certification, is required to establish trust on Indian data.

#### 5.5.2 Data Quality Assessment Tools

Automated data quality assessment using tools like F-UJI (Automated FAIR Data Assessment Tool), which evaluates core FAIR metrics based on the RDA FAIR Data Maturity Mode, should be implemented. [29,42] Assessment reports should identify missing variables, inconsistent coding, outliers suggesting data entry errors, metadata gaps, and interoperability issues. Data quality reports should be visible to downstream users but should not prevent sharing. A dataset with acknowledged quality gaps is more ethically sound than selective reporting where gaps are hidden.

#### 5.5.3 Academic Credit for Data Auditors and Peer Reviewers

Data auditing and peer review should be recognized as scholarly work eligible for promotion credit. A researcher who verifies another researcher's dataset, identifies quality issues, and documents findings in a "Data Audit Report" should receive API credit equivalent to that granted for a published commentary or technical note. Such recognition should incentivize rigorous data preparation by original collectors, encourage peer participation in quality assurance activities, and normalize quality-focused scholarship within academic evaluation systems.

#### 5.5.4 Transparent Methodology and Sensitivity Analyses

Any statistical analysis or AI model derived from shared data should include a detailed description of all data transformations, imputation methods, and analytical choices; sensitivity analyses demonstrating results under varying assumptions; stratified reporting by relevant subgroups; and explicit caveats regarding data limitations. Ensuring transparency of methodological choices should discourage selective reporting and promote the adoption of robust, defensible analytical approaches.

**5.5.5 Tracking data provenance and transparency**

Paper proformas and offline spreadsheets are prone to manipulation. Study proformas should be digital with time logs, requiring data entry at the time of recruitment rather than at the end. Research electronic data capture (REDCap) is such a free web/server software that keeps all the trail of every single change to data, making it very difficult to hide post-collection changes. It also allows granular access control, a kind of need-to-know or need-to-edit basis.

**5.5.6 Specialty-Specific Mandated Minimum Datasets**

Data should be mandated to be reported in standardized consensual terminology at the outset by the ethical committees. Specialty-specific minimum datasets should be defined in alignment with international standards such as HL7 FHIR and OMOP, while remaining simplified for Indian realities. Registries and AI projects adopting these standards should receive priority in ICMR/ANRF funding and fast-track ethics approval. Over time, payers may require conformance to minimum dataset standards as a condition for empanelment or reimbursement.

# 6. REFORMING POSTGRADUATE MEDICAL THESES: FROM FRAGMENTED PROJECTS TO MULTICENTRIC REGISTRIES

**6.1 Current Problem**

The NMC mandates a thesis for every postgraduate medical student in India, even though only a small fraction of these projects lead to publications, interoperable datasets, or meaningful research.[31,32] Most produce fragmented, under-powered theses and siloed data rather than robust shared registries.

Data are frequently collected hastily and non-systematically with small convenience samples, major gaps, and limited attention to reproducible methods. The compulsory single-centre thesis is often treated as a formality by residents and sometimes by faculty. The original policy justification was that "doing" research teaches research interpretation, but weakly designed, under-powered single-centre theses do little to teach robust methodology or contribute usefully to the evidence base.[43] This represents an enormous opportunity cost: large volumes of clinician and patient time are spent on small, duplicative projects with minor variations across institutions, producing datasets that cannot be meaningfully pooled.

## 6.2 Nodal Multicentric Model

Older or research-active medical colleges with a track record of original publications and methodological capacity should be designated as nodal centres for postgraduate thesis work within a state or region. Instead of each resident pursuing an isolated topic, nodal centres would coordinate multicentric protocols that include several colleges in adjoining districts and multiple residents from the same recruitment year.

Sample size and logistics would be planned with built-in redundancy so that even if two or three of six to eight centres drop out, the remaining data still yield a meaningful, adequately powered study. This structure directly addresses data fragmentation. The same effort currently spread across many small projects would feed into a single larger dataset that can support robust analysis and future AI and machine learning applications.

The nodal centre, led by an experienced principal investigator, would coordinate protocol design with inputs from participating centres and guides. Quarterly or half-yearly online meetings would track recruitment, protocol adherence, and data quality, with decisions and proposed changes logged prospectively. Domain expertise from all centres would be pooled at the design stage, and common data dictionaries and variable definitions would be agreed upon upfront to facilitate later pooling and secondary analysis.

Participating smaller colleges would primarily focus on high-quality data collection and first-line verification, calibrated to their capacity. Thesis guides at these centres, including those with limited prior research experience, would have a more defined and feasible role: ensuring ethical conduct, supervising recruitment and documentation, and verifying a defined proportion of records through callbacks or case review. This reduces the expectation that every guide independently designs and runs a full-scale study, while still embedding them in the research workflow and allowing gradual capacity building.

## 6.3 Quality Control and Governance

To prevent undue dominance by the nodal centre and to safeguard data integrity, a second independent "quality check" centre should audit a random sample of data. This centre's sole role in the project would be quality assurance rather than recruitment, creating a double layered quality control system. Thesis guides at peripheral centres would verify a predefined sample of cases through direct contact or review of source records, with the external quality centre cross checking another random subset.

All multicentric thesis projects should operate under transparent governance. Meetings should be recorded, minutes shared, and protocol amendments documented with clear attribution. Authorship and credit should be governed by pre-specified rules, not by informal hierarchies or institutional prestige.[44] A pre-weighted scoring system could allocate points for major contributions across domains such as protocol design, data collection volume and quality, data management and analysis, and manuscript drafting. If a junior faculty member from a newer centre contributes disproportionately to protocol development, recruitment, or analysis, they should receive appropriately senior authorship and greater visibility, even when the nodal institution is more reputed. This both incentivizes genuine engagement and mitigates fears that nodal models will simply siphon credit from peripheral centres.

## 6.4 Capacity Building and AI Readiness

The nodal model should not reduce peripheral centres to "data collection shops."[106] Rather, the policy should mandate that nodal centres offer opportunities for interested faculty and residents from participating sites to join subgroups focused on protocol refinement, data cleaning, statistical analysis, or manuscript drafting. These activities can be structured as optional tracks or electives for those who wish to develop deeper research skills, without forcing every resident or guide to shoulder the same level of responsibility. Over time, this creates a pipeline of clinicians with genuine experience in multicentric study design, database management, and collaborative writing.

From the perspective of AI and machine learning and national registries, the proposed structure offers several advantages. Harmonised, multicentric datasets, prospectively designed with common data dictionaries and quality checks, are far better suited for downstream machine learning, risk modelling, and decision support tools than hundreds of small, heterogeneous spreadsheets generated from isolated theses. If state-level nodal groups focus on priority conditions, repeated thesis cohorts can progressively enrich longitudinal registries that can then underpin both conventional epidemiological analyses and AI and machine learning applications such as predictive models for treatment response or adverse events.

## 6.5 Sustainability and Evaluation Reform

In the current system, many studies are abruptly terminated when the resident graduates, even if sample size targets are not achieved. A simple reform would allow "relay theses," where a protocol that has not reached its target can be continued by the next resident under the same faculty guide, with explicit attribution of contributions for each cohort[111]. This reduces redundancy and encourages continuity in research questions, methods, and data infrastructure[112]. Similarly, two residents in the same batch, supervised by different faculty members, may share a single multicentric protocol, pooling their patient loads and reducing the individual burden while improving power and generalisability.

Thesis evaluation should shift its emphasis from nominal "completion" of a stand-alone project to the quality and integrity of the data contributed to a shared project. Instead of forcing every candidate to prove that their personal study is fully powered, the system can recognise contributions to larger registries or multicentric datasets, provided predefined criteria for data completeness, accuracy, and follow-up are met. For residents who cannot commit to full protocol design and analysis, satisfactory completion might be defined as high-quality data collection and documentation within a multicentric project.

At the faculty level, promotion and recruitment criteria should explicitly reward contributions to multicentric research and registry building, separate from single-centre case series or convenience sample theses. This includes credit for protocol leadership, coordination roles, data management, and cross-institutional collaboration. Institutions should establish local committees or research offices that actively facilitate such collaborations, resolve inter-departmental or inter-institutional conflicts, and ensure equitable credit distribution, rather than passively "allowing" them.

## 7. ADDRESSING IMPLEMENTATION BARRIERS

### 7.1 Legal and Regulatory Ambiguity

The DPDPA research exemption is ambiguous. The government should issue formal guidance specifying that academic AI development (non-commercial, without individual-level decisions) falls within Section 17(2)(b). ICMR and ANRF should establish indemnification policies: if a researcher follows prescribed data governance standards, the liability should be of the party at whose end the breach occurred. The third-party processors should, by law, required to have strict safeguards to idiotproof any potential data leakages.

### 7.2 Metric Gaming and Goodhart's Law

When you gamify a system, people try to game the system. The targets and objectives should be flexible. Numbers/quantities should be monitored and published only with quality and contextual relevance in mind, and if they are actually fulfilling the aim/goal/vision for which they were designed. The intent behind the objectives should be the focus rather than the objectives themselves, and the processes should be rewarded not just outcomes. Strategies include giving more weightage for data which is

- rated of higher quality in peer review
- is successfully reused (equivalent to citation of an article)
- adding a large database in one go from a centre rather than in parts from the same time frame
- adding to an existing database on same parameters by a multi-centric or cross-institutional collaboration
- adding longitudinal follow-up data on the same cohort later to enable better understanding of the course of a disease or intervention

Such strategies will disincentivize researchers from racking up points by uploading incomplete, duplicated, or irrelevant datasets (equivalent to publishing non-novel or fake case reports in non-index predatory journals to rack up publication count), and prevent fragmenting of a data set (to gather more points by uploading them separately). If "percentage of patients consenting to data sharing" becomes a metric, clinicians will selectively approach compliant populations, introducing severe selection bias, and can target vulnerable populations without proper informed consent. So, data quality metrics, diversity indices, and refusal rates as proxy for consent quality need to be measured instead. Obviously, the hacks researchers used to exploit loopholes will evolve with time. In the era of AI and fast evolution, the government measures to counter these should keep up the pace.

### 7.3 Central Repository Breach Risk

Hospitals' concerns regarding breach risks from central data repositories should be addressed by prioritizing federated learning architectures, where raw data never leaves hospital premises. Redundant, geographically distributed repositories with encrypted replication should be established. Researchers should access only derived datasets or query interfaces through federated analytics, rather than raw records. [45, 46]

### 7.4 Sustainability of Data Stewardship

Funding gaps for data stewardship between 3- to 5-year grant cycles should be addressed by allowing ANRF to allocate institutional overhead (10–15% of grant value) specifically for stewardship activities. Dedicated "Data Stewardship Grants" (₹50–200 lakhs for 3 years) should be launched to support hiring data curators. Fee-for-service models should be implemented, with external researchers paying modest fees (₹10,000–50,000 per data request) to sustain ongoing stewardship.

### 7.5 Scalability of Federated Learning

The challenge of stable participation in federated learning across 30,000+ medical practitioners and thousands of clinics should be addressed through hierarchical structures: Tier 1 (major teaching hospitals as central nodes), Tier 2 (district/state hospitals as regional hubs), and Tier 3 (PHCs/CHCs as leaf nodes that aggregate data locally before participation). Asynchronous and dynamic participation should be enabled. Implementation should begin with 5–10 major teaching hospitals to establish protocols before scaling.

### 7.6 Equity and Power Asymmetries

To prevent large teaching hospitals from dominating federated learning and marginalizing smaller clinics, Shapley value distribution should be used to assign high value to rare data contributions. 20–30% of leadership roles should be reserved for smaller institutions. Large hospitals should be paired with smaller ones for capacity building. Differential incentives should be provided, with smaller institutions receiving higher compute credits and priority funding to offset resource constraints.

## 8. FUNDING

None of the above proposed measures, even in combination, would be as effective as a single measure: significantly increasing funding. India's budget allocation for research and development has a percentage of gross domestic product has been less than 1%, which is significantly less than the USA, European countries, and China.[47] A similar pattern is reflected in the Indian private setup, in both the start-ups as well as established companies. This can no longer be acceptable in the era of AI, when the research can be the gateway to exponential progress. In a world rapidly changing due to AI, a significant investment at this stage can compensate for decades of disadvantages which India has faced. But the development of AI primarily needs good data, and collecting good data needs money. It has to be made a priority by the government at every stage and step. One cannot expect already overburdened government hospital departments to digitize their old analog case files faithfully without providing expert manpower and data entry operators for the same. Indian health care providers will continue to have significant resistance to adoption of electronic health records if they are forced on it without accounting for the increased time per consult, which proper documentation will require. A persistent complaint of tech collaborators is that clinicians with the potential for most data (in government centers with heavy patient loads) are often very busy to collect it. Transcribing natural interactions of doctors with patients also require cloud storage, high quality sensitive microphones, software, and manual review, all of which needs money. This problem cannot be resolved through temporary fixes like having contractual staff, projects, or untrained data operators. Permanent additional faculty is more likely to be motivated to gather correct and quality data in form of large longitudinal cohorts with follow ups, and to contribute to improving an institute's data infrastructure. In this crucial juncture, India must prioritize research, which requires data, which requires priority investment.

## 9. CONCLUSION: FROM HOARDING TO VALUATION

India's data fragmentation is not accidental: it is a rational response to misaligned incentives. Until data curation is recognized as research, rewarded as professional work, and embedded in institutional rankings and funding priorities, researchers will continue to rationally hoard data or donate it informally, leaving the AI ecosystem starved of diverse, high-quality inputs. The key recommendations discussed in this paper are summarised across incentives, infrastructure, and governance domains in Table 1.

The proposed framework realigns individual and institutional self-interest with public health goals. A researcher at a district hospital who contributes validated cases to a national registry gains: academic credit toward promotion; professional recognition; institutional prestige; financial incentives if algorithms trained on the data are commercialized; and infrastructure access.

Simultaneously, the system creates accountability. Transparent data use agreements, federated architectures, automated quality assessment tools, and structured peer review ensure that data are used ethically and extracted value flows back to contributors. The solution is not secrecy but transparency: visible quality reports, documented methodology, stratified reporting, and academic credit for rigorous auditing. When methodological gaps are visible and auditing is rewarded, data sharing becomes less risky than selective reporting.

This is a call for systematic reform of how we recognize, value, and incentivize data work within the academic and clinical enterprise. If India builds such a collaborative data ecosystem, it will produce AI and registries that genuinely reflect its population, inform evidence-based policy, and improve patient care across all communities and geographies. India now faces a clear choice. One path continues current practices, producing isolated publications and fragmented datasets with limited long-term value. The other treats clinical data as shared public infrastructure, enabling collaborative research, representative registries, and future-ready analytical tools. Recent initiatives such as the Government of India's One Nation One Subscription policy reflect growing recognition that access to knowledge should be treated as shared public infrastructure, rather than an institutional privilege. The tools, precedents, and regulatory space are available globally. What remains is political will and institutional leadership to implement this framework.

# REFERENCES

1. Indian Council of Medical Research. National ethical guidelines for biomedical and health research involving human participants [Internet]. New Delhi: Indian Council of Medical Research; 2017 [cited 2026 Apr 12]. Available from: https://www.indiascienceandtechnology.gov.in/sites/default/files/file-uploads/guidelineregulations/1527507675_ICMR_Ethical_Guidelines_2017.pdf

2. Moher D; Shamseer L; Clarke M; Ghersi D; Liberati A; Petticrew M; et al. Preferred reporting items for systematic review and meta-analysis protocols (PRISMA-P) 2015 statement. Syst Rev. 2015;20(2):148–60. doi:10.1186/2046-4053-4-1 PubMed PMID: 25554246.

3. Guidelines for Ethical Use of Leftover Anonymous Samples for Commercial Purpose [Internet]. 2024 [cited 2026 Apr 12]. Available from: https://www.icmr.gov.in/icmrobject/uploads/Guidelines/1732704229_guidelinesforethicaluse.pdf

4. Ministry of Law and Justice G of I. The Digital Personal Data Protection Act, 2023 [Internet]. New Delhi, India: Gazette of India; 2023 Aug. Available from: https://www.meity.gov.in › uploads › 2024/06

5. Department of Science and Technology. National Data Sharing and Accessibility Policy-2012 (NDSAP-2012). The Gazette of India [Internet]. New Delhi, India; 2012 Mar [cited 2026 Apr 12]. Available from: https://dst.gov.in/national-data-sharing-and-accessibility-policy-0

6. National Data Sharing and Accessibility Policy | Department Of Science & Technology [Internet]. [cited 2026 Apr 12]. Available from: https://dst.gov.in/national-data-sharing-and-accessibility-policy-0

7. Ministry of Law and Justice G of I. The Anusandhan National Research Foundation Act, 2023 [Internet]. India: Gazette of India; 2023 Aug 14. Available from: https://dst.gov.in/sites/default/files/NRF.pdf

8. Department of Biotechnology G of I. Biotech-PRIDE Guidelines (Promotion of Research and Innovation through Data Exchange) [Internet]. New Delhi; 2021 Jul [cited 2026 Apr 12]. Available from: https://dbt.gov.in/storage/media/publication/pride.pdf

9. Ministry of Health and Family Welfare G of I. Strategy for Artificial Intelligence in Healthcare for India (SAHI) [Internet]. New Delhi; 2026 Feb [cited 2026 Apr 12]. Available from: https://abdm.gov.in/sahi

10. JC D, JL R, DB G, A P, JW S, G J, et al. The "All of Us" Research Program. N Engl J Med. 2019 Aug 15;381(7):668–76. doi:10.1056/NEJMSR1809937 PubMed PMID: 31412182.

11. Department of Science and Technology G of I. National Supercomputing Mission Annual Report 2023–24 [Internet]. New Delhi; 2024 [cited 2026 Apr 12]. Available from: https://nsmindia.in/wp-content/uploads/2025/04/NSM-AR-book-design-02425_updated_11zon.pdf

12. Hripcsak G, Duke JD, Shah NH, Reich CG, Huser V, Schuemie MJ, et al. Observational Health Data Sciences and Informatics (OHDSI): Opportunities for Observational Researchers. Stud Health Technol Inform. 2015;216:574. doi:10.3233/978-1-61499-564-7-574 PubMed PMID: 26262116.

13. Vuokko R, Vakkuri A, Palojoki S. Systematized Nomenclature of Medicine–Clinical Terminology (SNOMED CT) Clinical Use Cases in the Context of Electronic Health Record Systems: Systematic Literature Review. JMIR Med Inform. 2023;11:e43750. doi:10.2196/43750 PubMed PMID: 36745498.

14. National Resource Centre for EHR Standards (NRCeS) CDP. Implementation Guide for Adoption of FHIR in ABDM and NHCX [Internet]. Pune, India; 2024 Sep [cited 2026 Apr 12]. Available from: https://www.nrces.in › download › files › pdf › I...

15. Garza MY, Rutherford M, Myneni S, Fenton S, Walden A, Topaloglu U, et al. Evaluating the Coverage of the HL7® FHIR® Standard to Support eSource Data Exchange Implementations for use in Multi-Site Clinical Research Studies. AMIA Annual Symposium Proceedings. 2021;2020:472. PubMed PMID: 33936420.

16. NITI Aayog G of I. National Health Stack: Strategy and approach [Internet]. 2018 [cited 2026 Apr 12]. Available from: https://abdm.gov.in › strapicms › uploads

17. Ward R, Pratt N, Hart G, Meyers I, Sullivan C, Luxan BG, et al. Cancer Clinical Academic Group [Internet]. 2025 [cited 2026 Apr 12]. Australian Health Data Evidence Network (AHDEN): Building a National Data Infrastructure for Standardised, Federated Health Data Research. Available from: Australian Health Data Evidence Network (AHDEN) Observational Health Data Sciences and Informatics (OHDSI) https://www.ohdsi.org › wp-content › uploads › 2025/10

18. New HealthX Sandbox 2.0 accelerates HealthTech Innovation in Singapore [Internet]. [cited 2026 Apr 12]. Available from: https://www.biospectrumasia.com/news/46/26069/new-healthx-sandbox-2-0-accelerates-healthtech-innovation-in-singapore.html

19. Enterprise Singapore. Guidelines on Data Standards (Terminology) to Support Interoperability of Healthcare System Records. Enterprise Singapore; 2025.

20. El Sabawy D, Feldman J, Pinto AD. The Connected Care for Canadians Act: an important step toward interoperability of health data. CMAJ : Canadian Medical Association Journal. 2024 Dec 9;196(42):E1385. doi:10.1503/CMAJ.241123 PubMed PMID: 39653400.

21. Goldstein MM, Pewen WF. The HIPAA Omnibus Rule: Implications for Public Health Policy and Practice. Public Health Reports. 2013;128(6):554. doi:10.1177/003335491312800615 PubMed PMID: 24179268.

22. What is GDPR, the EU's new data protection law? - GDPR.eu [Internet]. [cited 2026 Apr 12]. Available from: https://gdpr.eu/what-is-gdpr/

23. NATIONAL MEDICAL COMMISSION POST-GRADUATE MEDICAL EDUCATION BOARD POST-GRADUATE MEDICAL EDUCATION REGULATIONS-2023.

24. University Grants Commission (UGC) I. University Grants Commission Notification on API (Academic Performance Indicators) Regulations, 2013. 2013 Jun.

25. Federer LM, Lu YL, Joubert DJ, Welsh J, Brandys B. Biomedical Data Sharing and Reuse: Attitudes and Practices of Clinical and Scientific Research Staff. PLoS One. 2015 Jun 24;10(6):e0129506. doi:10.1371/journal.pone.0129506

26. Pisani E, AbouZahr C. Sharing health data: good intentions are not enough. Bull World Health Organ. 2010 Jun 1;88(6):462–6. doi:10.2471/BLT.09.074393

27. Ministry of Education G of I. India Rankings 2025: National Institutional Ranking Framework (NIRF) – Medical. 2025.

28. HL7 International. HL7 Publishes FHIR® Release 4 [Internet]. 2019 [cited 2026 Apr 9]. Available from: https://blog.hl7.org/hl7-publishes-fhir-release-4

29. Wilkinson MD, Dumontier M, Aalbersberg IjJ, Appleton G, Axton M, Baak A, et al. The FAIR Guiding Principles for scientific data management and stewardship. Scientific Data 2016 3:1. 2016 Mar 15;3(1):160018-. doi:10.1038/sdata.2016.18 PubMed PMID: 26978244.

30. Data Management and Sharing Policy | Grants & Funding [Internet]. [cited 2026 Apr 12]. Available from: https://grants.nih.gov/policy-and-compliance/policy-topics/sharing-policies/dms

31. Mahajan R, Saiyad S. Postgraduate Medical Education Regulations 2023: A Critical Review. Int J Appl Basic Med Res. 2024 Jan;14(1):1. doi:10.4103/IJABMR.IJABMR_23_24 PubMed PMID: 38504845.

32. Woods HB, Pinfield S. Incentivising research data sharing: A scoping review. Wellcome Open Research. F1000 Research Ltd; 2021. doi:10.12688/wellcomeopenres.17286.1

33. Gliklich RE, Dreyer NA, Leavy MB. Registries for Evaluating Patient Outcomes. AHRQ Publication. 2014;1:669. PubMed PMID: 24945055.

34. Stall S, Bilder G, Cannon M, Chue Hong N, Edmunds S, Erdmann CC, et al. Journal Production Guidance for Software and Data Citations. Scientific Data. Nature Research; 2023. doi:10.1038/s41597-023-02491-7 PubMed PMID: 37752153.

35. Burton A, Koers H, Manghi P, Stocker M, Fenner M, Aryani A, et al. The scholix framework for interoperability in data-literature information exchange. D-Lib Magazine. 2017;23(1–2). doi:10.1045/january2017-burton

36. Ghorbani A, Zou J. Data Shapley: Equitable Valuation of Data for Machine Learning. In: Chaudhuri K; Sugiyama M, editor. Proceedings of the 36th International Conference on Machine Learning (ICML 2019) [Internet]. Long Beach, California, USA: PMLR (Proceedings of Machine Learning Research); 2019 [cited 2026 Apr 9]. p. 2242–51. Available from: https://proceedings.mlr.press/v97/ghorbani19c.html

37. Kairouz P, McMahan HB, Avent B, Bellet A, Bennis M, Bhagoji AN, et al. Advances and Open Problems in Federated Learning. Foundations and Trends in Machine Learning. 2019 Dec 10;14(1–2):1–210. doi:10.1561/2200000083

38. Kuo TT, Kim HE, Ohno-Machado L. Blockchain distributed ledger technologies for biomedical and health care applications. Journal of the American Medical Informatics Association. Oxford University Press; 2017. p. 1211–20. doi:10.1093/jamia/ocx068 PubMed PMID: 29016974.

39. Azaria A, Ekblaw A, Vieira T, Lippman A. MedRec: Using blockchain for medical data access and permission management. In: Proceedings - 2016 2nd International Conference on Open and Big Data, OBD 2016. Institute of Electrical and Electronics Engineers Inc.; 2016. p. 25–30. doi:10.1109/OBD.2016.11

40. Zhang P, White J, Schmidt DC, Lenz G, Rosenbloom ST. FHIRChain: Applying Blockchain to Securely and Scalably Share Clinical Data. Comput Struct Biotechnol J. 2018 Jan 1;16:267–78. doi:10.1016/j.csbj.2018.07.004

41. Devaraju A, Huber R. F-UJI - An Automated FAIR Data Assessment Tool [Internet]. doi:10.5281/ZENODO.4063720

42. Ioannidis JPA. Why Most Published Research Findings Are False. PLoS Med. 2005 Jul 23;2(8):e124. doi:10.1371/JOURNAL.PMED.0020124 PubMed PMID: 16060722.

43. International Committee of Medical Journal Editors (ICMJE). ICMJE Recommendations: Preparing for Submission [Internet]. 2024 [cited 2026 Apr 9]. Defining the Role of Authors and Contributors. Available from: https://www.icmje.org/recommendations/browse/roles-and-responsibilities/defining-the-role-of-authors-and-contributors.html

44. Rieke N, Hancox J, Li W, Milletarì F, Roth HR, Albarqouni S, et al. The future of digital health with federated learning. npj Digital Medicine 2020 3:1. 2020 Sep 14;3(1):119-. doi:10.1038/s41746-020-00323-1

45. Ji M, Xu G, Ge J, Li M. Efficient Core-selecting Incentive Mechanism for Data Sharing in Federated Learning [Internet]. 2023 Sep 21 [cited 2026 Apr 9]. Available from: https://arxiv.org/pdf/2309.11722

46. Ministry of Science and Technology G of I. Press Information Bureau (PIB), Government of India [Internet]. [cited 2026 Apr 12]. Parliament Question: R&D Investment in India. Available from: https://www.pib.gov.in/PressReleasePage.aspx?PRID=2153547

**Table 1. Summary of Proposed Reforms**

| Domain | Core Problem | Key Recommendations | Expected Outcome |
|---|---|---|---|
| **Incentives** | Data work is invisible in academic and professional evaluation | • Recognise curated datasets, registry contributions, and data audits as scholarly output<br>• Include data contributions in promotion criteria, funding decisions, and institutional rankings | Researchers are rewarded for data stewardship, not just publications |
| **Infrastructure** | Fragmented, non-interoperable data systems | • Define specialty-specific minimum datasets<br>• Build interoperable registries and shared platforms<br>• Use federated architectures where centralisation is risky | High-quality, reusable datasets that support registries, research, and analytics |
| **Governance** | Legal uncertainty and fear of misuse discourage sharing | • Issue clear regulatory guidance and safe harbours<br>• Use transparent data use agreements<br>• Implement quality assessment and fair attribution mechanisms | Reduced risk, greater trust, and ethical reuse of clinical data |